\documentclass[10pt,twocolumn,letterpaper]{article}

\usepackage[pagenumbers]{cvpr} 
\usepackage{tabularx}
\usepackage{array} 
\usepackage{amsfonts}
\usepackage{amssymb}
\usepackage{booktabs}
\usepackage{float}
\usepackage{multirow}

\definecolor{cvprblue}{rgb}{0.21,0.49,0.74}
\usepackage[pagebackref,breaklinks,colorlinks,allcolors=cvprblue]{hyperref}

\def\paperID{5473} 
\def\confName{CVPR}
\def\confYear{2025}

\title{Local Features Meet Stochastic Anonymization: Revolutionizing Privacy-Preserving Face Recognition for Black-Box Models}

\author{
Yuanwei Liu$^{1,2}$\quad
Chengyu Jia$^{1,2}$\quad
Ruqi Xiao$^{1,2}$\quad
Xuemei Jia$^{1,2}$\quad
Hui Wei$^{1,2}$\\
Kui Jiang$^{3}$\quad
Zheng Wang$^{1,2}$\thanks{Corresponding authors.}\quad
\\
\normalsize{$^1$National Engineering Research Center for Multimedia Software, Institute of Artificial Intelligence, School of} \\ 
\normalsize{Computer Science, Wuhan University  \quad ${^2}$Hubei Key Laboratory of Multimedia and Network Communication Engineering, }\\
\normalsize{Wuhan University, China \quad ${^3}$School of Computer Science and Technology, Harbin Institute of Technology, Harbin, China.}
}

\begin{document}
\maketitle

\begin{abstract}
The task of privacy-preserving face recognition (PPFR) currently faces two major unsolved challenges: (1) existing methods are typically effective only on specific face recognition models and struggle to generalize to black-box face recognition models; (2) current methods employ data-driven reversible representation encoding for privacy protection, making them susceptible to adversarial learning and reconstruction of the original image. We observe that face recognition models primarily rely on local features (\emph{e.g.}, face contour, skin texture, and so on) for identification. Thus, by disrupting global features while enhancing local features, we achieve effective recognition even in black-box environments. Additionally, to prevent adversarial models from learning and reversing the anonymization process, we adopt an adversarial learning-based approach with irreversible stochastic injection to ensure the stochastic nature of the anonymization. Experimental results demonstrate that our method achieves an average recognition accuracy of 94.21\% on black-box models, outperforming existing methods in both privacy protection and anti-reconstruction capabilities.


\end{abstract}

\begin{figure}[tbp]
  \centering
   \includegraphics[width=\linewidth]{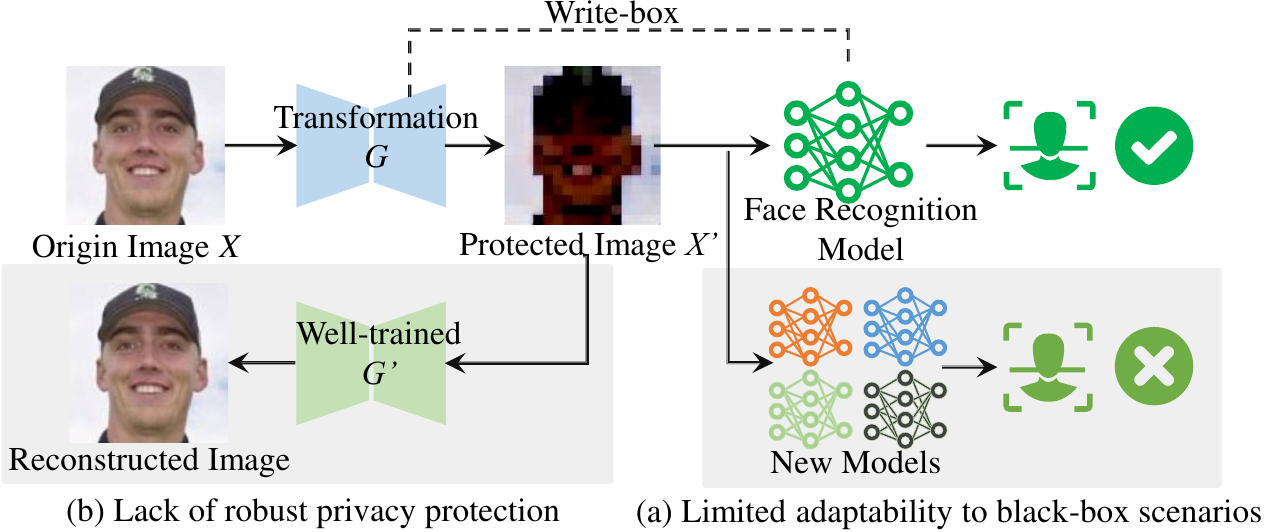}
   \caption{\textbf{SOTA methods face two primary challenges.} (a) \textbf{Lack of robust privacy protection}: SOTA methods are vulnerable to reconstruction models $G'$ that can learn the encoding scheme and recover the original image. (b) \textbf{Limited adaptability to black-box scenarios}: SOTA methods are tailored to specific face recognition models, limiting their effectiveness on general-purpose face recognition models.} 
   \label{fig:paradigm}
\end{figure}

\section{Introduction}
\label{sec:intro}
Privacy-Preserving Face Recognition (PPFR) is gaining growing significance in the digital society, particularly in light of the extensive adoption of face information~\cite{r1,r2,r3}.
The goal of PPFR is to retain face recognition capabilities while utilizing technical methods to maximize user privacy protection. Specifically, face images uploaded by users can be encrypted, preventing identity thieves from recognizing them while still enabling reliable identity verification.

Current state-of-the-art PPFR methods are mainly categorized into cryptographic-based~\cite{r7,r4,r9,r10,r11} and transformation-based~\cite{r14,r17,r22,r23,r24,r27} approaches. Cryptographic-based methods use advanced encryption algorithms to keep data encrypted during transmission and processing. However, their high computational cost and the complexity of managing private keys have limited widespread use. In contrast, transformation-based methods alter images or feature vectors to retain critical distinguishing features while obscuring visual personal information. These characteristics make transformation-based methods more practical and widely to be adopted.

In transformation-based methods, given a face image $X$, a transformation $G$ generates a protected image $X' = G(X)$ that aims to achieve anonymity while preserving recognition accuracy. However, maintaining high recognition accuracy often compromises the anonymization of $X'$, making it vulnerable to re-identification with minor modifications. All existing methods encounter two significant, often overlooked challenges: (1) \textit{Limited adaptability to black-box scenarios.} The transformation $G$ modifies image features but is customized for specific face recognition models. To achieve optimal performance with new models, retraining is required, which limits flexibility. (2) \textit{Lack of robust privacy protection.} The mapping $X \rightarrow X'$ via $G$ is deterministic, allowing attackers to potentially reconstruct identity information from $X'$ using reconstruction networks. If a sufficient number of $(X, X')$ pairs are available, the transformation can be reverse-engineered, rendering the system vulnerable to reconstruction attacks~\cite{r30,r31,r32}.


To address these challenges, we draw inspiration from the ``holistic effect" in cognitive science. Numerous studies~\cite{jacobs2019comparing,geirhos2018imagenet,richler2014meta} indicate that humans rely primarily on global features and rapidly form an overall impression for face recognition, while machine learning models achieve recognition through hierarchical extraction and analysis of numerous local features. We observe that by disrupting global features while preserving local features, it is possible to reduce visual informativeness while retaining key components essential for model-based discrimination.

This paper proposes a novel privacy-preserving face recognition (PPFR) method that maintains high accuracy on black-box and universal face recognition models without the need to train a dedicated model. Unlike traditional methods that use data-driven, reversible representation encoding, our approach introduces an irreversible stochastic injection mechanism through adversarial learning. By constructing a non-deterministic stochastic mapping in the transformation from the input image $X$ to the anonymized image $X'$, our method effectively resists existing reconstruction attacks, significantly enhancing anonymization performance.

In the data preprocessing process, we leverage differences between the human visual system and machine models in information processing. Global features perceptible to the human eye are pruned, retaining only the essential local details required for face recognition. Additionally, we employ a stochastic nonlinear algorithm to fill the pruned regions, further enhancing the level of anonymization and increasing the difficulty of reconstruction.

We then fine-tune the feature distribution in high-dimensional space by adding noise to the preprocessed images. In this process, we design multi-scale temporal noise and introduce an entropy-increasing regularization term in the loss function. This enhances the randomness in the $X \to X'$ mapping, thereby obstructing reconstruction networks from learning the mapping patterns. To ensure robust transferability in black-box face recognition models, we utilize a momentum-driven multi-model ensemble optimization strategy. Experimental results demonstrate that our method achieves an average accuracy of 94.21\% on black-box general face recognition models, effectively preventing reconstruction while offering superior privacy protection compared to existing methods. 

Our main contributions are summarized as follows:



\begin{itemize}
    \item We propose the first privacy-preserving face recognition method that operates effectively on black-box models without requiring dedicated retraining, ensuring compatibility with general face recognition systems.
    \item Our approach uniquely applies targeted adversarial training in the image domain, incorporating multi-scale temporal noise and entropy regularization to achieve a non-deterministic mapping, significantly enhancing resilience against reconstruction attacks. This innovation greatly enhances resilience against reconstruction attacks while achieving an average accuracy of 94.21\% on black-box models, outperforming existing methods in both privacy protection and robustness.
\end{itemize}

\section{Related Work}
\subsection{Face Recognition Models Compatibility}
Transformation-based approaches are the mainstream methods for privacy-preserving face recognition tasks. Primarily, techniques such as noise injection based on specific distributions~\cite{r14,r15,r16,r17}, linear mixing with perturbation images~\cite{r18}, and adversarial training~\cite{r19,r20,r15,r21} are commonly used to obscure the original visual content. However, they often degrade recognition performance due to their indiscriminate application of perturbations. Other simpler anonymization approaches~\cite{r22}, such as blurring, pixelation, or local face replacement, while maintaining accuracy for certain visual features, fail to offer robust privacy protection. More sophisticated techniques~\cite{r23,r24,r25,r26,r27}, which involve transforming face images into frequency-domain representations followed by feature filtering or reorganization, have shown promise in striking a balance between privacy preservation and recognition accuracy. Nevertheless, these techniques typically generate specific protective representations incompatible with general-purpose face recognition models. Thus, they necessitate concurrently training a specialized face recognition model to ensure both privacy protection and high accuracy. This not only escalates implementation costs but also limits the model's generalizability and transferability.

In contrast, we employ the momentum iterative method in combination with multi-model ensemble optimization, enabling high accuracy on existing face recognition black-box models without requiring model re-deployment. This strategy effectively tackles the challenge of incompatibility with face recognition models faced by existing methods and offers scalability.

\subsection{Resistance to Reconstruction Attacks}
To effectively protect privacy, face recognition systems must both generate privacy-preserving representations in real-time and be resilient to adversarial attacks that attempt to reconstruct the original image from these representations. Reconstruction attacks can be broadly categorized into two types: optimization-based methods~\cite{r28,r29}, which iteratively adjust the input image pixels to minimize the discrepancy between the reconstructed and target features, and learning-based methods~\cite{r30,r31,r32,r33,r34}, which leverage large datasets to train neural networks that directly reconstruct images from the privacy-preserving representations, achieving higher reconstruction quality at lower attack cost.

Among existing privacy protection techniques, some~\cite{r34} effectively defend against reconstruction attacks from feature representations to images through plug-in mechanisms, yet they fail to address privacy concerns at the level of the original image. Others~\cite{r26,r23,r24} attempt to balance image anonymization with the irreversibility of privacy protection using random channel selection or feature reorganization techniques. However, these perturbative operations do not fundamentally alter the recognizable features of the image. This fixed mapping between the original image and key features remains vulnerable to attacks, as adversaries, in the presence of rich training data, can learn to bypass perturbations and successfully reconstruct the face image.

Our method diverges from existing privacy protection methods that focus on the extraction of feature channels. We innovatively introduce targeted adversarial training directly in the image domain, combining multi-scale temporal adversarial optimization with entropy-increasing regularization to implement a stochastic mapping mechanism. This mechanism effectively nullifies reconstruction attacks, offering a more robust and comprehensive solution to privacy protection.

\begin{figure*}[t]
  \centering
    \includegraphics[width=\textwidth]{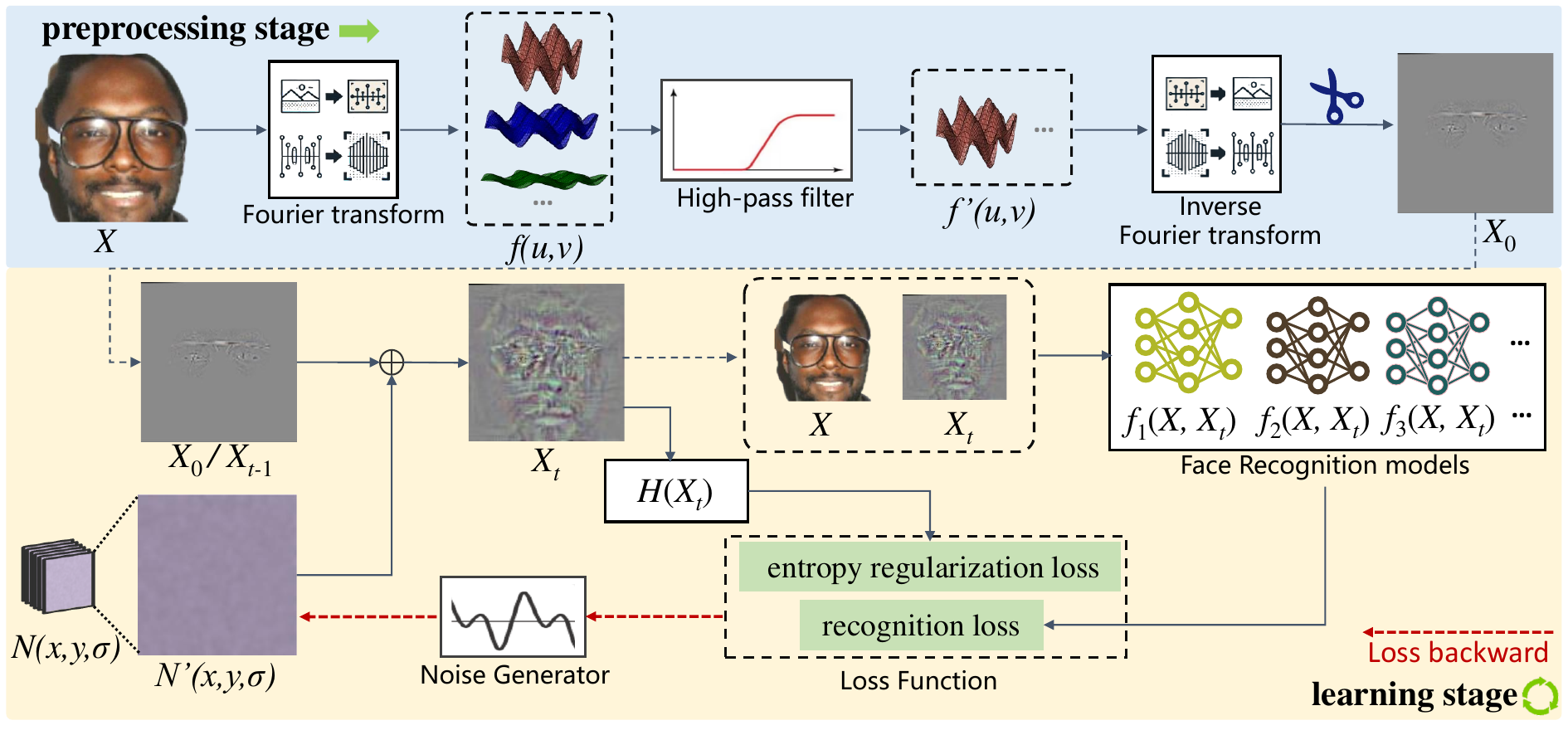}
  \caption{\textbf{The overview of our method.} The pipeline consists of two stages: preprocessing and learning. 1) In the preprocessing stage, we apply a Fourier transform to image $X$ to obtain its frequency representation $f(u, v)$ and preserve high-frequency components $f'(u, v)$ through high-pass filtering. After inverse transformation, pruning, and stochastic nonlinear filling, the initial image $X_0$ is obtained. 2) In the learning stage, $X_0$ serves as the starting image, with multi-scale temporal noise $N'(x, y, \sigma)$ added iteratively to generate $X_t$. Both $X_t$ and the original image $X$ are input into an ensemble model to compute recognition loss, combined with an entropy-regularization term $H(X_t)$ to optimize the generated noise. After multiple iterations, the final anonymized image $X'$ is produced.}
  \label{fig:our anonymization}
\end{figure*}

\section{Our PPFR Methodology}
\label{sec:method}
Sec.~\ref{sec:Motivation} presents the motivation and overall framework of our proposed method. Sec.~\ref{sec:3.2} describes the image preprocessing technique designed to effectively separate global and local features. Sec.~\ref{sec:3.3} and Sec.~\ref{sec:3.4} provide details on the incorporation of randomization within the optimization process. Finally, Sec.~\ref{sec:3.5} focuses on the construction of the optimizer and loss function to ensure high accuracy within black-box face recognition models.

\subsection{Motivation}
\label{sec:Motivation}
Our approach is inspired by the distinct mechanisms through which the human visual system and machine learning models process facial information. The human visual system primarily relies on prominent global features, such as the overall face shape, contour, and color distribution, to recognize faces. This holistic processing mechanism enables rapid identification based on a cohesive visual impression of face features. In contrast, machine learning models employ a hierarchical extraction of local features, leveraging deep feature extraction techniques to capture fine-grained structural details, including face edges, textures, and geometric relationships among face landmarks. This characteristic allows models to excel in distinguishing subtle differences in face attributes, thereby achieving high-precision face recognition.


Motivated by the differential reliance on feature types between the human visual system and machine learning models, we propose a novel privacy-preserving method, with the overall pipeline illustrated in Fig.~\ref{fig:our anonymization}. By obscuring global features and retaining only local features, our approach anonymizes face images for human observers while allowing effective recognition by machine models. We further introduce adversarial noise into the preserved local features to fine-tune their distribution, enhancing their distinctiveness within the model’s high-dimensional feature space and activating discriminative channels, thereby maintaining accuracy even in black-box models. Additionally, substantial randomness is incorporated into the optimization process, making the anonymization mapping resistant to adversarial learning and effectively mitigating reconstruction attacks to ensure robust privacy protection.

\subsection{Fine-Grained Feature Extraction and Filtering}
\label{sec:3.2}
In the preprocessing stage, our objective is to remove the globally recognizable features easily discernible by the human eye to achieve anonymization. Due to the high sensitivity of the Fourier transform to spatial variations, it effectively captures regions of rapid changes, such as edges and fine textures, which contain rich detail information.

Thus, given an image $X$ with a width $W$ and height $H$, we decompose the image using Fourier transform, where each sinusoidal component represents a specific pattern:
\begin{equation}
f(u, v)=\mathcal{F}(X)=\sum_{x=0}^{H-1} \sum_{y=0}^{W-1} v(x, y) \cdot e^{-2 \pi i\left(\frac{u x}{H}+\frac{v y}{W}\right)}.
\end{equation}
The fine details, \textit{e.g.} edges and textures, manifest as higher-order variations, characterized by shorter oscillation periods, which are encoded in the Fourier coefficients $F(u, v)$.

To eliminate the macroscopic global information and preserve key face details and local structures, we designed a high-pass filter,
\begin{equation}
H(u,v) = \mathbb{I} \left( \sqrt{u^2 + v^2} > \omega_C \right),
\end{equation}
where larger frequencies $u$ and $v$ correspond to smaller-scale, rapid variations. By employing a frequency threshold $\omega_C$, we effectively isolate the local microstructures in the image. 

The separated components are then transformed back into the spatial domain using the inverse Fourier transform to extract fine-scale image $X'$:

\begin{equation}
\begin{aligned}
f'(u,v)&=f(u,v)\cdot H(u,v), \\
X_0&=\mathcal{F}^{-1}(F'(u,v)) \\
&=\frac{1}{MN} \sum_{x=0}^{H-1} \sum_{y=0}^{W-1} f'(u,v)\cdot e^{2 \pi i\left(\frac{u x}{H}+\frac{v y}{W}\right)}.
\end{aligned}
\end{equation}

To further enhance the anonymization process and Randomization effect, we apply an additional transformation to the image. Specifically, while preserving the most detailed and model-sensitive regions (such as the eyes), the remaining areas are filled using a stochastic nonlinear function $R(x,y)$,
\begin{equation}
R(x, y) \sim \mathcal{N}(\mu(x, y), \sigma(x, y) \cdot \kappa).
\end{equation}
This step ensures that only the most critical local features are retained, while the randomized treatment of other regions strengthens the anonymization and mitigates the risk of reconstructing the original image.

The function $R(x,y)$ is generated using a Gaussian distribution based on the image’s color mean $\mu(x, y)$ and standard deviation $\sigma(x, y)$. Specifically, $\mathcal{N}(x,y)$ represents a normal distribution where the mix is the mean and $\sigma(x, y) \cdot \kappa $ is the adjusted standard deviation, with $\kappa$ acting as a scaling factor for the standard deviation to control the spread of the stochastic values.

\subsection{Multi-Scale Temporal Adversarial Optimization}
\label{sec:3.3}
Most existing face anonymization techniques rely on training a generative network to process images. Although some studies claim that additional operations can prevent reconstruction attacks, the fundamental mechanism behind these methods remains a reversible representation encoding $X'=G(X,\theta_G)$. As a result, with a sufficiently large dataset for training a reconstruction network, the risk of reconstruction remains inherent.

To increase reconstruction difficulty and enhance randomization, we design a multi-scale temporal noise approach. Specifically, using a temporal seed $\sigma$, we generate noise image $N(x,y,\sigma)$ at each level through random gradient vectors $h(i,j,\sigma)$ and a smoothing function $\phi(x-i,y-i)$.,
\begin{equation}
    N(x,y,\sigma)=\sum_{i,j}h(i,j,\sigma) \cdot \phi(x-i,y-i).
\end{equation}
The noise image order $k$ governs the increasing frequency, such that higher orders introduce finer details with larger frequencies $\tau$, while the persistence $\varepsilon$ reduces and the corresponding weights gradually diminish. By superimposing multiple orders of noise images, we generate a final complex noise structure:
\begin{equation}
N'(x,y,\sigma)=\sum_{k=0}^{n} \varepsilon ^k\cdot N(\tau ^k\cdot \frac{x}{\lambda} ,\tau ^k\cdot \frac{y}{\lambda},\sigma).
\end{equation}
This multi-scale architecture, coupled with the frequency multiplication mechanism, introduces subtle perturbations at various scales, significantly increasing the complexity of the noise patterns and disrupting the reconstruction process.

\subsection{Entropy-Enhanced Loss Regularization}
\label{sec:3.4}

Entropy in an image is a statistical concept used to measure the amount of information or complexity, representing the intricacy of pixel distribution. We observe that anonymization methods that are harder to reconstruct tend to produce protected images with higher entropy. Further analysis reveals that this phenomenon arises because common image reconstruction models, such as U-net~\cite{miccai/RonnebergerFB15}, primarily rely on convolutional operations to extract spatial features, 
\begin{equation}\label{convolutional}
    y(i, j)=\sum_{m=0}^{k-1} \sum_{n=0}^{k-1} W(m, n) \cdot I(i+m, j+n),
\end{equation}
which in turn depends on spatial consistency within the image. Images with higher entropy typically contain more information, characterized by greater differences between adjacent pixels and more fragmented, irregular textures, which disrupt spatial consistency.

Therefore, we incorporate an Entropy-Enhanced Loss Regularization\eqref{entropy} into the loss function to guide the pixel value distribution toward randomness,
\begin{equation}\label{entropy}
    H(X)=-\sum_{i=1}^{n} p(X_i)\log{p(X_i)}.
\end{equation}
It disrupts the stability and consistency of the features extracted by convolutional neural networks. This, in turn, significantly increases the reconstruction difficulty for the reconstruction model.


\subsection{Construction of Optimizer and Loss Function}
\label{sec:3.5}
In this section, we focus on the construction of the optimizer and the loss function, specifically detailing how to leverage the previously discussed concepts and appropriate techniques to ensure that highly anonymized images exhibit high transferability, ultimately achieving effective recognition by black-box face recognition models.

For the initial image $X$, we aim to iteratively optimize it through an optimizer to ultimately obtain the protected image $X'$. To accelerate the optimization process and avoid local minima, we leverage accumulated historical gradients, enhancing the transferability of the optimization results. Simultaneously, by performing a weighted average of predictions from multiple ensemble models, we effectively reduce the model variance without significantly increasing bias, thereby improving the generalization capability of the results. Based on these two principles, and incorporating the multi-scale temporal optimization noise discussed in Sec.~\ref{sec:3.3}, we constructed the following optimizer:

\begin{equation}
    X_{t+1}=X_t+N'(x,y,\sigma )\cdot\alpha\cdot(\mu \cdot g_t+\frac{\nabla _XL_{total}(X_t)}{||\nabla _XL_{total}(X_t)||_p}).
\end{equation}

Here, $N'(x,y,\sigma)$ denotes the multi-scale temporal noise introduced in Sec.~\ref{sec:3.3}, $\alpha$ is the learning rate, $\mu$ is the momentum coefficient, and $g$ is the historical gradients.

The overall loss $L_{total}$ consists of two components: the recognition loss and the entropy regularization loss,
\begin{equation}
L_{total}=L_{fr}+L_{entropy}=\sum_{i=1}^{n}\beta_i \cdot f_i(X,X_t) +\gamma \cdot\mathbb{E}(H(X_t)).
\end{equation}

Let $f$ denote each surrogate face recognition model. The model extracts features from two images $X$ and $X_t$, then computes the cosine similarity between their face features. Based on a predefined threshold, the model determines whether the two input images correspond to the same individual. We directly use the output cosine similarity as part of the recognition loss, with $\beta$ representing the parameters associated with each surrogate model. Additionally, the function $H$ refers to the entropy regularization term introduced in Sec.~\ref{sec:3.4}, with $\gamma$ representing its corresponding parameter.


\section{Experiments}
\label{sec:exp}

\subsection{Experimental Setup}
\label{sec:4.1}
\paragraph{Datasets.} In this experiment, we employ the \textsc{LFW}~\cite{LFWTech}, \textsc{CelebA}~\cite{liu2015deep}, \textsc{AgeDB}~\cite{moschoglou2017agedb}, \textsc{CPLFW}~\cite{zheng2018cross}, and \textsc{CALFW}~\cite{zheng2017cross} datasets to evaluate the face recognition accuracy of our method. Specifically, \textsc{LFW} contains over 13,000 face images, each representing a unique individual; \textsc{CelebA} includes 202,599 celebrity face images of 10,177 distinct individuals; \textsc{AgeDB} contains 16,488 images of 100 individuals across various age groups; \textsc{CPLFW} consists of 3,000 face pairs emphasizing pose variation; and \textsc{CALFW} comprises 3,000 cross-age face pairs.


\paragraph{Face Recognition Models.} We leverage six prominent face recognition models, including ArcFace~\cite{deng2019arcface}, Mobilenet~\cite{howard2017mobilenets}, Resnet~\cite{he2016deep}, IR~\cite{li2007illumination}, FaceNet~\cite{schroff2015facenet}, and CosFace~\cite{wang2018additive}. These models are employed to optimize the anonymization process and assess the effectiveness of various PPFR methods in maintaining recognition performance. All models achieved an accuracy exceeding 99\% on the \textsc{LFW} dataset.



\paragraph{Compared Methods.} Four existing advanced PPFR methods are considered: (1) \textbf{Duetface}~\cite{r27}, appling channel segmentation in the frequency domain combined with client-server collaboration; (2) \textbf{Proface}~\cite{r22}, employing a Siamese network to integrate original and blurred face images; (3) \textbf{PartialFace}~\cite{r24}, randomly selecting subsets of high-frequency components from the original image; and (4) \textbf{MinusFace}~\cite{r23}, creating visually uninformative images through feature subtraction. We compare our method with these approaches across three critical aspects: recognition performance, visual anonymization, and robustness against reconstruction attacks, providing a comprehensive assessment of its privacy-preserving effectiveness.


\paragraph{Evaluated Metrics.} (1) For the \textbf{visual anonymization} task in Sec.~\ref{sec:4.2}, we employ the Structural Similarity Index Measure (SSIM) to evaluate the anonymization performance. SSIM, rooted in the structural perception of the human visual system, quantifies the structural similarity between the anonymized and original images, thereby providing a robust measure of anonymization effectiveness. (2) For the \textbf{accurate identification task} in Sec.~\ref{sec:4.3}, we leverage the thresholding mechanism of a face recognition model to verify whether the anonymized and original images correspond to the same individual. Identification accuracy is computed over a large-scale dataset to quantitatively assess the retention of identity recognizability post-anonymization. (3) For \textbf{resilience to reconstruction attacks} task in Sec.~\ref{sec:4.4}, we extend our evaluation by incorporating the Peak Signal-to-Noise Ratio (PSNR) alongside SSIM to assess the quality of reconstructed images. PSNR quantitatively characterizes the level of noise in the reconstructed image, providing an additional metric to gauge the robustness against reconstruction attacks.

\begin{figure}[t] 
    \centering
    \includegraphics[width=1.0\linewidth]{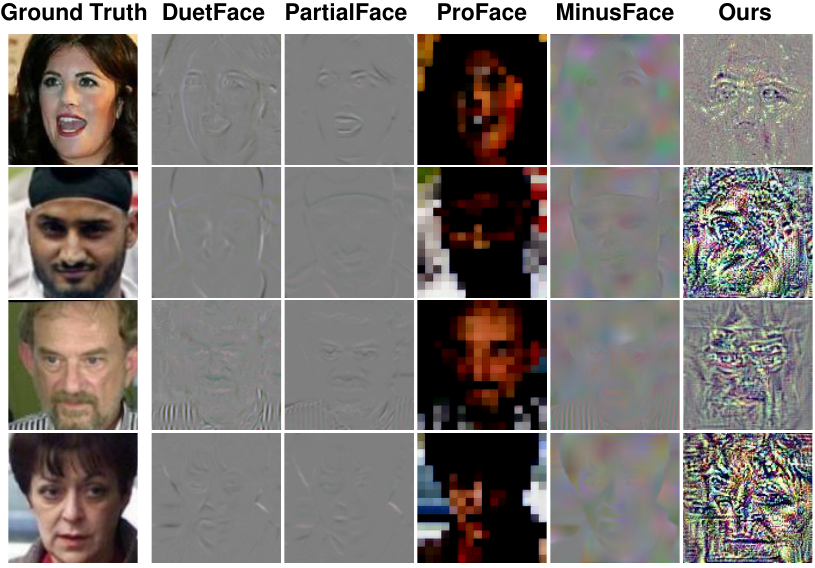} 
    \caption{\textbf{Subjective comparison of SOTA methods and ours.} We present the ground truth alongside anonymized results from existing methods, including DuetFace~\cite{r27}, PartialFace~\cite{r24}, ProFace~\cite{r22}, and MinusFace~\cite{r23}. While these methods retain the overall facial structure, our approach completely removes global features.}
    \label{fig:anonymization comparison}
\end{figure}

\subsection{Anonymization Effect}
\label{sec:4.2}
In this study, we conduct a comprehensive analysis of the first objective in the PPFR task—\textbf{visual anonymization}. As illustrated in Fig.~\ref{fig:anonymization comparison}, we compare our proposed method with the state-of-the-art (SOTA) approaches discussed in Sec.~\ref{sec:4.1}. These SOTA methods employ specific transformation techniques to isolate certain visual features but still retain the overall structure of the original face image, allowing a degree of visual recognizability. In contrast, our method completely eliminates global features, retains only local features, and incorporates stochastic noise for further obfuscation. Comparative results indicate that images generated by our approach are entirely unrecognizable to the human eye. Quantitative analysis in Tab.~\ref{tab:table1} demonstrates that our method consistently outperforms existing SOTA methods across all metrics used to assess image similarity.

\begin{table}[t]
\setlength{\tabcolsep}{2pt} 
\renewcommand\arraystretch{1.1}
\resizebox{1.0\linewidth}{!}{
\begin{tabular}{lcccccc}
\toprule
Method & Venue& \textsc{LFW} & \textsc{CelebA} & \textsc{AgeDB} & \textsc{CPLFW} & \textsc{CALFW} \\

\midrule
DuetFace~\cite{r27} & MM'22 & 0.37 & 0.32 & 0.31 & 0.37 & 0.31 \\
ProFace~\cite{r22} & MM'22  & 0.22  & 0.30  & 0.29  & 0.31  & 0.24  \\
PartialFace~\cite{r24} & ICCV'23 & 0.35  & 0.31  & 0.32  & 0.35  & 0.30 \\
MinusFace~\cite{r23} & CVPR'24 & 0.34  & 0.29  & 0.30 & 0.36  & 0.29  \\
\textbf{Ours} & Ours & \textbf{0.11}  & \textbf{0.07} & \textbf{0.09}  & \textbf{0.12}  & \textbf{0.15}  \\
\bottomrule
\end{tabular}
}
\caption{\textbf{Quantitative results of anonymization effectiveness between SOTA methods and our approach.}The similarity between anonymized and original images is evaluated using the SSIM. By calculating the SSIM value between the anonymized and original images, we measure the effectiveness of anonymization. A lower SSIM value indicates a lower visual similarity between the two images, signifying a stronger anonymization effect. }
\label{tab:table1} 
\end{table}

As shown in Fig.~\ref{fig:our anonymization}, we randomly selected several images for anonymization to illustrate the visual effects. The anonymized images exhibit significant variations in saturation, contrast, and color distribution, highlighting that our method’s stochastic generation mechanism introduces substantial diversity into the anonymized image dataset.

\begin{figure*}
  \centering
  \includegraphics[width=\textwidth]{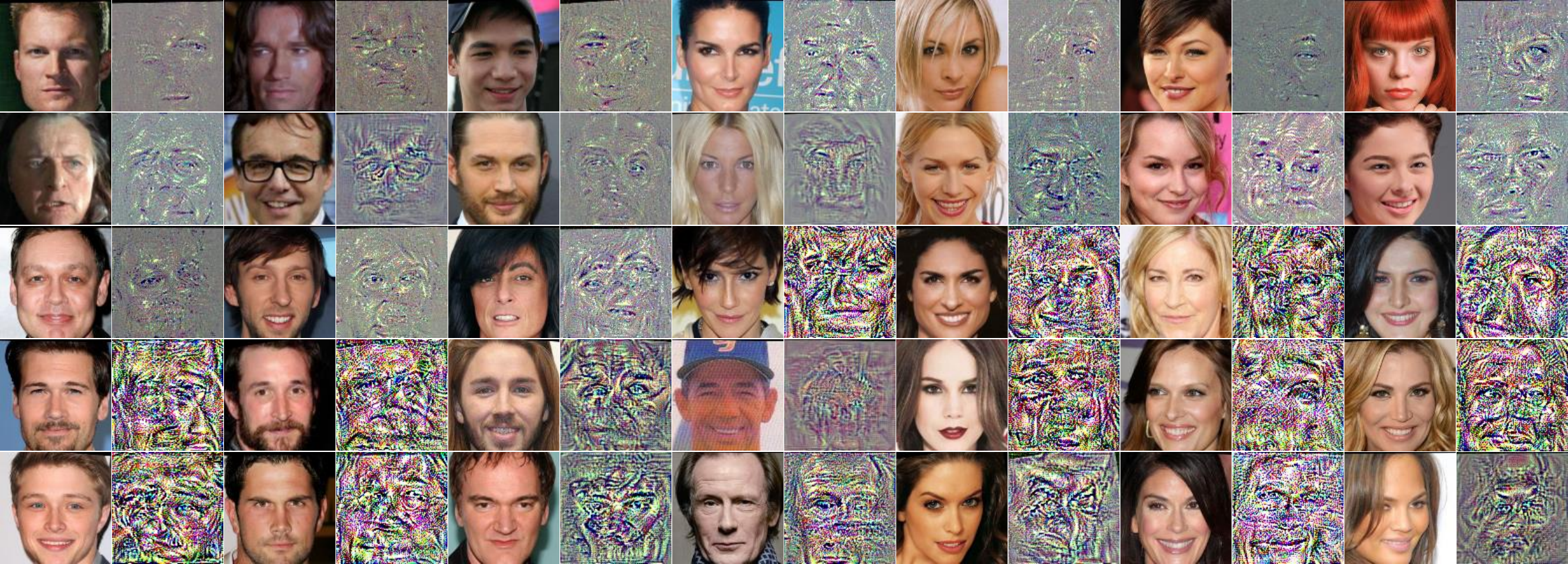}
  \caption{\textbf{The visualization of the anonymization effect of our method.} We selected a diverse set of samples across various datasets to demonstrate the anonymization effectiveness of our method. The figure showcases a range of face samples varying in age, gender, and expression, capturing a wide spectrum of individual characteristics. For each pair, the left side shows the original face, while the right side displays the image processed by our anonymization technique. The anonymized images exhibit variations in saturation, contrast, and color distribution, reflecting the diversity and richness of the samples.}
  \label{fig:our anonymization}
\end{figure*}

\subsection{Black-box Model Accuracy}
\label{sec:4.3}
Subsequently, we conduct a detailed analysis of the second objective in the PPFR task—\textbf{accurate identification}. Existing anonymization methods alter critical features required for face recognition during image processing, which, we hypothesize, limits their effectiveness to only the face recognition models they were trained on, potentially failing under black-box settings with general-purpose face recognition models. Tab.~\ref{tab:table2} validates this hypothesis: under black-box settings, the identification accuracy of these methods remains below 2\%, with over half of the methods achieving 0\% accuracy.

\begin{table}[t]
\centering
\setlength{\tabcolsep}{2pt} 
\renewcommand\arraystretch{1.1}
\resizebox{1.0\linewidth}{!}{
\begin{tabular}{l c c c c c c c c}
\toprule
\textbf{Method}  & \textbf{ArcFace} & \textbf{Moblienet} & \textbf{Resnet} & \textbf{IR} & \textbf{FaceNet} & \textbf{CosFace} \\
\midrule
DuetFace~\cite{r27}  & \ \ \ \ \ \ \ 0\% & \ \ \ \ \ \ \ 0\% & \ \ \ \ \ \ \ 0\% & \ \ \ \ \ \ \ 0\% & \ \ 0.50\% & \ \ \ \ \ \ \ 0\% \\
ProFace~\cite{r22}  & \ \ 1.75\% & \ \ 0.75\% & \ \ \ \ \ \ \ 0\% & \ \ \ \ \ \ \ 0\% & \ \ 1.00\% & \ \ 0.50\% \\
PartialFace~\cite{r24}  & \ \ \ \ \ \ \ 0\% & \ \ 0.25\% & \ \ 0.75\% & \ \ \ \ \ \ \ 0\% & \ \ \ \ \ \ \  0\% & \ \ 0.75\% \\
MinusFace~\cite{r23}  & \ \ \ \ \ \ \  0\% &\ \ \ \ \ \ \ 0\% & \ \ \ \ \ \ \ 0\% & \ \ 0.25\% & \ \ 1.50\% & \ \ \ \ \ \ \ 0\% \\
\textbf{Ours} & \textbf{94.75\%} & \textbf{98.50\%} & \textbf{98.75\%} & \textbf{96.50\%} & \textbf{92.00\%} & \textbf{99.75\%} \\
\bottomrule
\end{tabular}
}
\caption{\textbf{Accuracy comparison between SOTA methods and our approach on black-box face recognition models. }Testing on \textsc{LFW} reveals that SOTA methods show minimal effectiveness on face models in a black-box setting, while our approach achieves an average accuracy of 96.71\%.}
\label{tab:table2}
\end{table}

\begin{table*}[t]
\setlength{\tabcolsep}{2pt} 
\renewcommand\arraystretch{1.1}
\resizebox{1.0\linewidth}{!}{
\begin{tabular}{lccccccccccc}
\toprule
\multirow{2}{*}{\textbf{Method}} &\multirow{2}{*}{\textbf{Venue}}&  \multicolumn{2}{c}{\textsc{LFW}} & \multicolumn{2}{c}{\textsc{CelebA}} & \multicolumn{2}{c}{\textsc{AgeDB}} & \multicolumn{2}{c}{\textsc{CPLFW}} & \multicolumn{2}{c}{\textsc{CALFW}}\\
\cmidrule(lr){3-4} \cmidrule(lr){5-6} \cmidrule(lr){7-8}\cmidrule(lr){9-10}\cmidrule(lr){11-12}
\multirow{2}{*}{} & \multirow{2}{*}{}  & SSIM($\downarrow$) & PSNR($\downarrow$) & SSIM($\downarrow$) & PSNR($\downarrow$)  & SSIM($\downarrow$) & PSNR($\downarrow$)& SSIM($\downarrow$) & PSNR($\downarrow$)& SSIM($\downarrow$) & PSNR($\downarrow$)\\
\midrule
DuetFace~\cite{r27} & MM'22 & 0.63 & 16.27 & 0.61 & 16.37 & 0.55 & 16.02 & 0.67 & 20.37 & 0.56 & 15.72 \\
ProFace~\cite{r22} & MM'22 & 0.79 & 19.28 & 0.65 & 18.52 & 0.62 & 17.64 & 0.66 & 17.27 & 0.75 & 15.95 \\
PartialFace~\cite{r24} & ICCV'23 & 0.62 & 16.91 & 0.69 & 17.58 & 0.55 & 14.83 & 0.60 & 15.23 & 0.59 & 16.78 \\
MinusFace~\cite{r23} & CVPR'24 & 0.52 & 16.41 & 0.62 & 15.78 & 0.48 & 14.42 & 0.53 & 16.24 & 0.50 & 14.65 \\ 
\textbf{Ours} & - & \textbf{0.45} & \textbf{12.20} & \textbf{0.44} & \textbf{15.53} & \textbf{0.43} & \textbf{12.80} & \textbf{0.42} & \textbf{13.30} & \textbf{0.48} & \textbf{14.20} \\
\bottomrule
\end{tabular}
}
\caption{\textbf{Objective Comparison of Reconstruction Resistance Between SOTA Methods and Our Approach.} Similarity between the reconstructed and original images is evaluated using SSIM and PSNR metrics. }
\label{tab:table4} 
\end{table*}

Our anonymization method preserves the local features essential for face recognition and has been evaluated across five major face datasets and six mainstream face recognition models. Experimental results indicate that under black-box settings, our method achieves an average identification accuracy of 94.21\%, demonstrating its effectiveness with general-purpose face recognition models.

\begin{table}[t]
\centering
\setlength{\tabcolsep}{2pt} 
\renewcommand\arraystretch{1.1}
\resizebox{1.0\linewidth}{!}{
\begin{tabular}{lccccc}
\toprule
\textbf{FR Model} & \textsc{LFW} & \textsc{CelebA} & \textsc{AgeDB} & \textsc{CPLFW} & \textsc{CALFW} \\
\midrule
ArcFace~\cite{deng2019arcface} & 94.75\% & 84.86\%& 90.13\% & 86.12\% & 91.98\% \\
Mobilenet~\cite{howard2017mobilenets} & 98.50\% & 90.93\% & 98.99\% & 96.40\% & 99.50\% \\ 
Resnet~\cite{he2016deep} & 98.75\% & 93.45\% & 99.24\% & 96.40\% & 99.50\% \\ 
IR~\cite{li2007illumination} & 96.50\% & 85.39\% & 98.48\% & 91.00\% & 96.49\% \\  
FaceNet~\cite{schroff2015facenet} & 92.00\%& 85.98\%& 92.15\%& 85.35\%& 92.98\% \\
CosFace~\cite{wang2018additive} & 99.75\%& 95.72\%& 99.75\% & 95.89\%& 99.50\% \\
\bottomrule
\end{tabular}
}
\caption{\textbf{Accuracy performance of our method across different face recognition models and datasets.} Our method achieves consistently high performance across different datasets and face models, with an average accuracy of 94.21\%.}
\end{table}

\subsection{Protection against Recovery}
\label{sec:4.4}
Finally, we conduct an in-depth analysis of the third objective in the PPFR task—\textbf{resilience to reconstruction attacks}. Reconstruction models are trained on a large set of $(X, X')$ pairs to learn the mapping from the anonymized image $X'$ back to the original image $X$, with the goal of recovering the original image. We utilize a U-net architecture~\cite{miccai/RonnebergerFB15} and the BUPT dataset, containing 1.3 million face images, as the training set to train a dedicated reconstruction model for each privacy-preserving method discussed.

Most existing advanced methods rely on reversible representation encoding, where anonymized images $X' = G(X)$ are generated through a specific model $G$. This reversible representation encoding allows reconstruction models to learn and exploit the encoding patterns effectively. In contrast, our method employs a stochastic mapping approach, resulting in a diverse set of anonymized samples without a consistent mapping pattern, thus rendering effective reconstruction infeasible. The quantitative results in Tab.~\ref{tab:table4} further substantiate our method’s robustness against reconstruction attacks by demonstrating significantly lower image similarity metrics.

\begin{figure}[t] 
    \centering
    \includegraphics[width=0.47\textwidth]{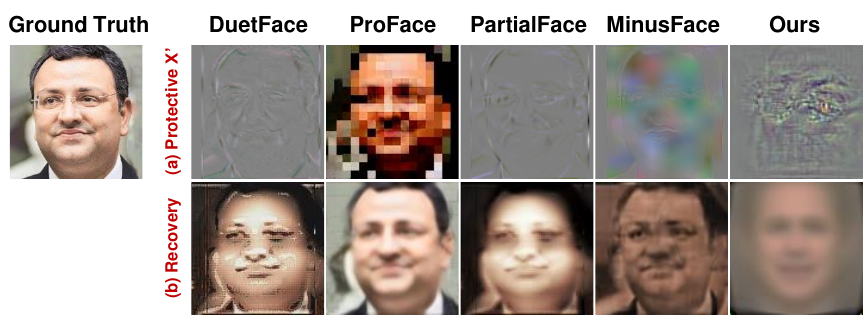} 
    \caption{\textbf{Subjective comparison of resilience to reconstruction attack between SOTA and our methods.} Specifically, (a) shows the anonymization results of SOTA methods and ours, while (b) displays the respective recovered images obtained from (a).}
    \label{fig:reconstruction comparison}
\end{figure}

As illustrated in Fig.~\ref{fig:reconstruction comparison}, although certain methods have made efforts to counteract reconstruction attacks and claim a level of robustness, their anonymized images remain identifiable when subjected to reconstruction models trained on large datasets. In contrast, our method prevents the learning of anonymization patterns, resulting in reconstructed images that only produce an average face devoid of any identifiable features.

\subsection{Ablation study}

We conducted ablation experiments to evaluate the impact of the model ensemble on maintaining high recognition accuracy for anonymized images under black-box settings. Specifically, we assembled ensembles of 1, 2, 3, and 4 models and tested their accuracy on a black-box face recognition model, as shown in Tab.~\ref{tab:table5}. To improve accuracy, the models in the ensemble need to exhibit sufficient diversity, meaning they should interpret the data differently, resulting in varied error patterns across samples. Our findings indicate that a minimum of three models is required to maintain a high recognition accuracy, aligning with the odd-number effect in ensemble theory. As the number of models increases, the marginal effect diminishes; each additional model contributes progressively less to performance improvement. The information gained from adding more models is likely low, as certain data characteristics are already well captured by the initial models in the ensemble.

\begin{table}[t]
\centering
\setlength{\tabcolsep}{2pt}
\renewcommand\arraystretch{1.1}
\resizebox{1.0\linewidth}{!}{
\begin{tabular}{c c c c c c c c}
\toprule
\textbf{\#FR} & \textbf{ArcFace} & \textbf{Mobilenet} & \textbf{Resnet} & \textbf{IR} & \textbf{FaceNet} & \textbf{CosFace}  \\
\midrule
1 & 14.00\% & 55.25\% & 18.00\% & 76.75\% & 30.00\% & 83.00\% \\
2 & 49.50\% & 79.75\% & 48.50\% & 78.00\% & 37.00\% & 88.50\% \\ 
3 & 94.75\% & \textbf{98.50\%} & 98.75\% & \textbf{96.50\%} & 61.00\% & 95.25\%   \\ 
4 & \textbf{96.25\%} & 97.75\% & \textbf{99.00\%} & 92.50\% & \textbf{92.00\%} & \textbf{99.75\%} \\  
\bottomrule
\end{tabular}
}
\caption{\textbf{The effect of model ensemble size on the accuracy of black-box face recognition models.} In each column, the model indicated by the column header is excluded from the optimization process for that model, ensuring a fully black-box setup.}
\label{tab:table5}
\end{table}

Additionally, we performed ablation experiments to examine the contribution of each method module proposed in Sec.~\ref{sec:method} to anonymization effectiveness and reconstruction resistance. We denote these modules as $\mathcal{F}$ for detail extraction, $\mathcal{C}$ for the stochastic nonlinear padding function, and $\mathcal{P}$ for multi-scale temporal noise construction and entropy-increasing regularization. These modules contribute to eliminating global features and extracting local features while continuously introducing random variables to establish stochastic mappings. Thus, these modules play a pivotal role in enhancing both anonymization effectiveness and reconstruction resistance, as corroborated by the results in Tab.~\ref{tab:table6}.
\renewcommand\arraystretch{1.1}
\begin{table}[H]
\centering
\begin{tabular}{l c c c c }
\toprule
\multirow{2}{*}{\textbf{Method}} & \multicolumn{2}{c}{\textbf{Anonymization}} & \multicolumn{2}{c}{\textbf{Construction}} \\
\cmidrule(lr){2-3} \cmidrule(lr){4-5}
\multirow{2}{*}{} &  SSIM($\downarrow$) & PSNR($\downarrow$)  & SSIM($\downarrow$) & PSNR($\downarrow$)  \\
\midrule
\textit{w/o} $\mathcal{F}$  & 0.39 & 21.73  & 0.62 & 16.75 \\ 
\textit{w/o} $\mathcal{C}$  & 0.09 & 25.58  & 0.60 & 14.98 \\
\textit{w/o} $\mathcal{P}$ & 0.13 & 18.32  & 0.57 & 16.82 \\
\textbf{Ours}  & \textbf{0.11} & \textbf{ 17.75}  & \textbf{0.45} & \textbf{12.20} \\  
\bottomrule
\end{tabular}
\caption{\textbf{Ablation studies on modules $\mathcal{F}$, $\mathcal{C}$, and $\mathcal{P}$.} We compare methods with and without these modules by evaluating the protected and reconstructed images using SSIM, and PSNR to assess anonymization effectiveness and reconstruction resistance. }
\label{tab:table6}
\end{table}

\section{Conclusion}
This paper introduces a novel approach to Privacy-Preserving Face Recognition (PPFR) that effectively addresses the critical challenges of limited adaptability to black-box models and vulnerability to reconstruction attacks. By preserving local features essential for recognition, obscuring global features, and incorporating stochastic anonymization through adversarial noise and entropy-enhanced regularization, our method achieves a robust balance between privacy protection and recognition accuracy. Our framework provides a scalable, secure, and practical solution for PPFR, paving the way for broader applications in privacy-sensitive face recognition scenarios.

\label{sec:conclusion}

{
    \small
    \bibliographystyle{ieeenat_fullname}
    \bibliography{main}
}


\end{document}


\maketitle

\section{Detailed Experimental Setups}
\label{detailed exp}

This section provides a detailed explanation of our experimental setup and the training specifics of the reconstruction models.
\subsection{Implementation Details}

\paragraph{High-Pass Filtering with Fourier Transform.} To process the input image \( X \), we use Fourier transform to decompose it into sinusoidal components, where each represents a specific spatial pattern:
\[
F(u, v) = \mathcal{F}(X) = \sum_{x=0}^{H-1} \sum_{y=0}^{W-1} X(x, y) \cdot e^{-2 \pi i \left( \frac{u x}{H} + \frac{v y}{W} \right)}.
\]
High-frequency details, such as edges and textures, are captured in the Fourier coefficients \( F(u, v) \), which represent rapid variations at smaller spatial scales. To preserve these details and suppress global structures, we apply a high-pass filter:
\[
H(u, v) = \mathbb{I} \left( \sqrt{u^2 + v^2} > \omega_C \right),
\]
where \( \mathbb{I} \) is an indicator function. The frequency threshold \( \omega_C \) is dynamically determined based on the characteristics of each image, ensuring adaptability to varying input content.

Finally, the filtered components are transformed back into the spatial domain using the inverse Fourier transform, yielding a fine-scale image \( X' \) that retains edges and textures while removing global patterns.

\paragraph{Stochastic Nonlinear Padding.} After filtering, pruned regions were filled with a stochastic nonlinear padding function defined as:
\[
R(x, y) \sim \mathcal{N}(\mu(x, y), \sigma(x, y) \cdot \kappa),
\]
where \( \mu(x, y) \) and \( \sigma(x, y) \) represent the local mean and standard deviation of pixel intensities, and \( \kappa = 1.5 \) controls variability. This dynamic padding ensures local consistency while introducing randomness.

\paragraph{Optimization and Loss Function.} The anonymized image \( X_t \) is optimized iteratively using gradient descent. The total loss function combines recognition loss and entropy regularization:
\[
L_{\text{total}} = L_{\text{fr}} + L_{\text{entropy}} = \sum_{i=1}^{n} \beta_i \cdot f_i(X, X_t) + \gamma \cdot \mathbb{E}(H(X_t)).
\]
\begin{itemize}
    \item The recognition loss term \( L_{\text{fr}} \) evaluates embedding alignment for surrogate models using cosine similarity. The weight \( \beta_i = 200 \) is assigned to this term to emphasize its importance.
    \item The entropy term \( L_{\text{entropy}} \), weighted by \( \gamma = 50 \), encourages randomness in the transformed image.
\end{itemize}

Momentum (\( \mu = 1.0 \)) is used to stabilize the optimization, and learning rates were dynamically adjusted as \( \alpha = 100 \cdot \epsilon / \text{iters} \), with \( \epsilon = 3 \) and \(\text{iters} = 100\).

\paragraph{Experimental Setup.} All experiments were conducted on a Linux server running Ubuntu 22.04.5 LTS, equipped with two NVIDIA RTX 3090 GPUs (24 GB VRAM each), using PyTorch 1.12 with CUDA 11.7.
\textbf{}

\subsection{Reconstruction Model}
\paragraph{Model Architecture.} For the reconstruction model, we adopted a UNet-based architecture~\cite{miccai/RonnebergerFB15}, which is widely recognized for its effectiveness in image-to-image translation tasks. The UNet model employs an encoder-decoder structure with skip connections, enabling the preservation of high-resolution spatial information while performing hierarchical feature extraction. This architecture is particularly suited for tasks that require fine-grained image reconstruction, as it balances local detail preservation with global structural consistency. 

\paragraph{Input and Feature Flow.} Specifically, the reconstruction model takes as input the anonymized images and aims to reconstruct the original images. The network operates on RGB image inputs with a resolution of 112×112. The skip connections facilitate the direct flow of low-level feature information from the encoder to the decoder, mitigating information loss and enhancing reconstruction accuracy. 

\heading{Training Setup.} The model was trained using a dataset of paired images, where the anonymized images served as input and the original images as the ground truth. We utilized the Adam optimizer with a learning rate of \(1 \times 10^{-4}\) and incorporated a learning rate scheduler (ReduceLROnPlateau) to adaptively adjust the learning rate based on validation loss trends. Training stability and efficiency were further enhanced through the use of mixed-precision training, leveraging the \texttt{torch.cuda.amp} module.

\subsection{Other Setups}
\paragraph{Comparison Methods.} Representative methods in face recognition privacy protection include PEEP~\cite{peep}, Cloak~\cite{cloak}, InstaHide~\cite{r18}, PPFD-FD~\cite{r25}, DCTDP~\cite{r26}, DuetFace~\cite{r27}, PartialFace~\cite{r24}, ProFace~\cite{r22}, AdvFace~\cite{r34}, ProFaceC~\cite{profacec}, and MinusFace~\cite{r23}. Among these, we selected DuetFace, PartialFace, ProFace, and MinusFace for comparison with our approach, as they represent state-of-the-art (SOTA) approaches that simultaneously address the requirements of face anonymization and resistance to reconstruction while maintaining high face recognition accuracy. This comparison effectively highlights the competitiveness of our method.
AdvFace was excluded because it focuses on feature vector perturbation rather than image-level protection, differing fundamentally from our objectives. Other methods were also excluded due to limitations such as insufficient advancement, poor performance in recognition accuracy and anonymization, or lack of consideration for reconstruction attacks.

\paragraph{Evaluation Metrics.} We selected SSIM to evaluate anonymization effectiveness and both Structural Similarity Index (SSIM)~\cite{wang2004image} and Peak Signal-to-Noise Ratio (PSNR) to assess reconstruction performance. Specifically, SSIM quantifies the structural similarity between the original and processed images based on the overall perception of the human visual system, with an emphasis on assessing the effectiveness of concealing facial contours and fine texture details. This aligns more closely with human sensitivity to facial anonymization and reconstruction quality. In contrast, PSNR focuses on pixel-level errors, compensating for SSIM's insensitivity to subtle pixel variations. This provides an additional measure of pixel-level similarity for evaluating reconstruction performance. However, since human recognition of facial features is not affected by pixel-level variations, PSNR is unsuitable as a metric for anonymization effectiveness.

\section{Further Ablation Study}
\label{ablation}
In this section, we provide additional experiments to supplement the ablation studies in the main text.
\begin{figure}[htbp] 
    \centering
    \includegraphics[width=1.0\linewidth]{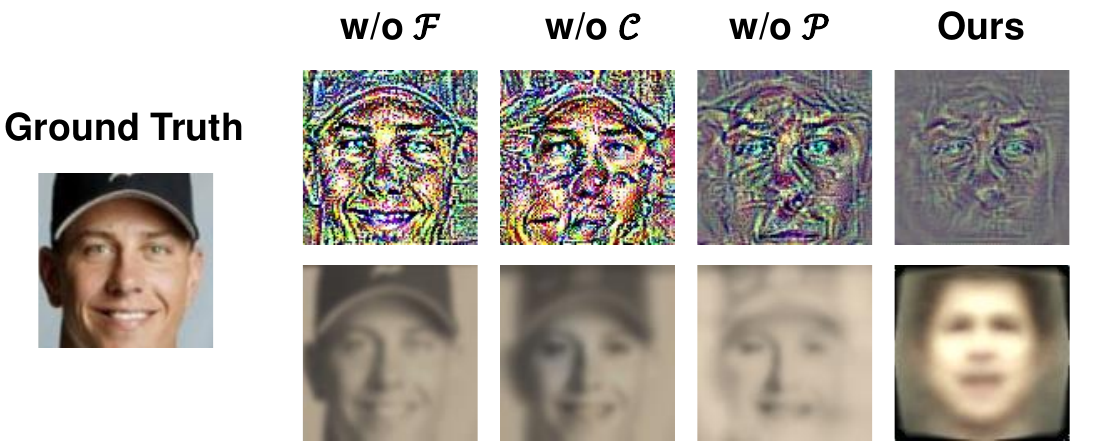} 
    \caption{\textbf{The visual results of anonymization and reconstruction after module ablation.} The first row illustrates the anonymization results, while the second row shows the reconstruction results. The ablation of modules results in suboptimal anonymization performance, whereas the full module configuration achieves significantly improved anonymization. In terms of reconstruction, the ablated modules produce high-fidelity results, while the full module configuration yields an "average face."
}
    \label{fig:ablation}
\end{figure}

\subsection{Further Ablation of Modules}
In the main text, we discuss the significance of the $\mathcal{F}$, $\mathcal{C}$ and $\mathcal{P}$ modules in achieving effective anonymization and reconstruction resistance. Ablation studies are conducted to validate the contributions of these modules to improving both anonymization and reconstruction resistance. Fig.~\ref{fig:ablation} provides a visual comparison to subjectively demonstrate the effects of ablating these three modules. As observed, the absence of individual modules leads to a noticeable decline in anonymization performance, with faint facial features still recognizable, making reconstruction possible. In contrast, the full module configuration achieves superior anonymization, effectively removing all facial features, thus preventing reconstruction.

\subsection{Recovery Result of SOTAs}
Several prior studies have claimed that their proposed methods are resistant to recovery attacks. However, as demonstrated in Fig. 5, our results show that these methods remain partially reversible. Based on our analysis in the main text, we attribute this issue to the inherent learnability of data-driven, reversible representation encoding. To further understand the discrepancies in experimental outcomes, we systematically analyze the causes from the following two perspectives:

\paragraph{Data Perspective.} Some methods rely on relatively small or limited datasets for training, restricting the recovery model's ability to learn transformation patterns effectively. In contrast, an attacker utilizing larger and more diverse datasets can significantly enhance recovery performance. Our experiments confirm this: as the dataset size increases, the quality of reconstructed images improves substantially. Specifically, using the BUPT dataset~\cite{bupt}, which contains 1.3 million images, the recovery model can generate images that reveal identifiable facial features. However, under the same experimental conditions, our method consistently produces only "average faces," further demonstrating its inherent unlearnability due to the strong randomness and lack of structural regularity in its transformations.

\begin{figure}[htbp] 
    \centering
    \includegraphics[width=1.0\linewidth]{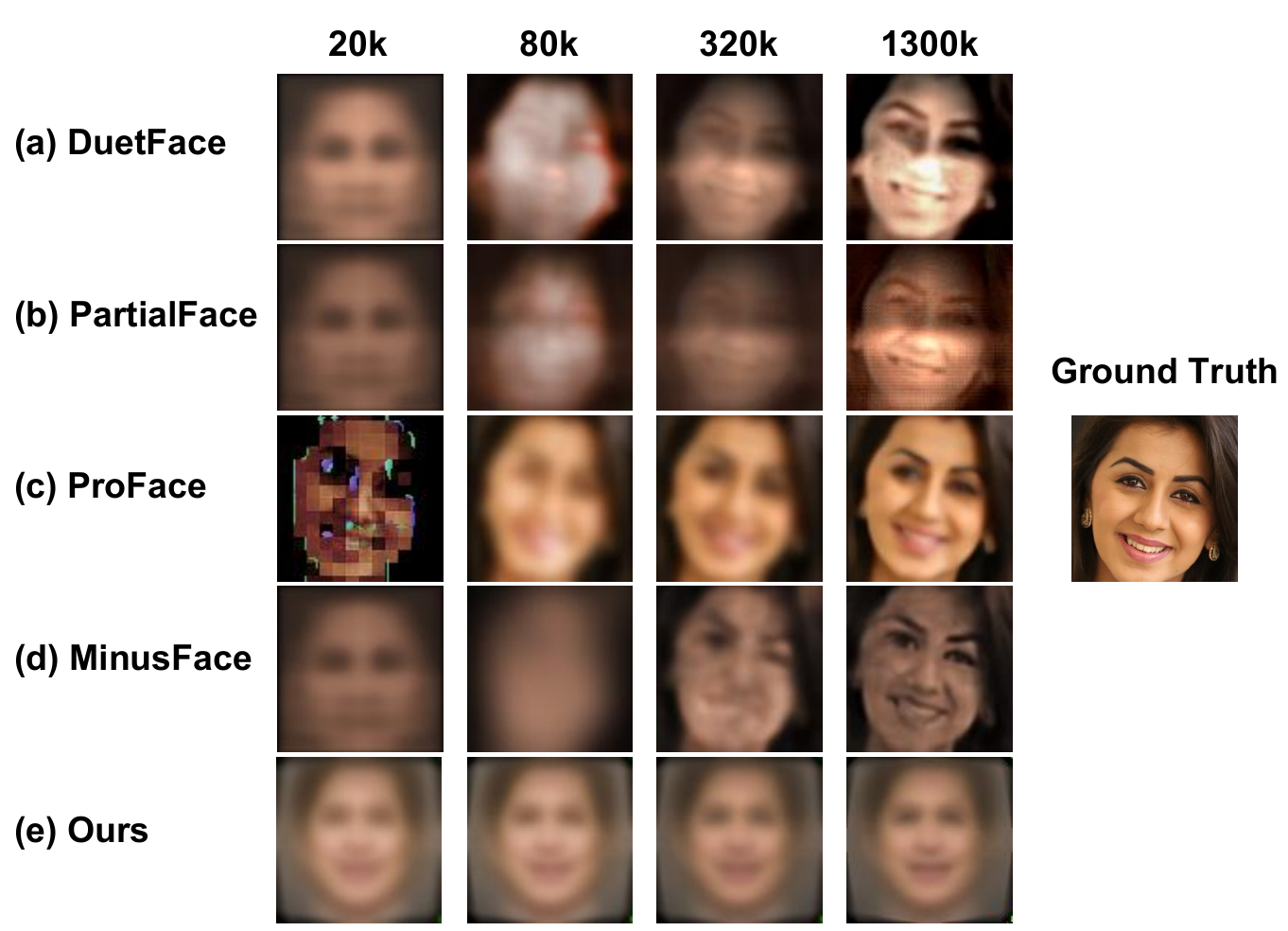} 
    \caption{\textbf{Impact of training data size on reconstruction models.} We trained reconstruction models using datasets of varying sizes and evaluated their ability to reconstruct protected images generated by different methods. The results demonstrate that other methods exhibit some resistance to reconstruction when the models are trained on smaller datasets. However, when the models are trained on the full BUPT dataset, the protected images from these methods can still be successfully reconstructed. In contrast, our method consistently produces a stable "average face," regardless of the training data size.
}
    \label{fig:recovery_size}
\end{figure}

\paragraph{Training Strategy Perspective.} The choice of training strategy plays a critical role in determining the performance of recovery models. Existing methods may suffer from instability or limited learning capacity due to suboptimal choices of hyperparameters, such as overly large learning rates or batch sizes. In our approach, we not only adopt appropriate learning rates and batch sizes but also incorporate multiple loss functions, such as perceptual loss and mean squared error (MSE) loss. This combination improves training stability and significantly enhances the recovery model’s ability to learn transformation patterns, resulting in superior recovery performance.

\section{Additional Example Images}
\label{example}

\begin{figure}[hbpt] 
    \centering
    \includegraphics[width=1.0\linewidth]{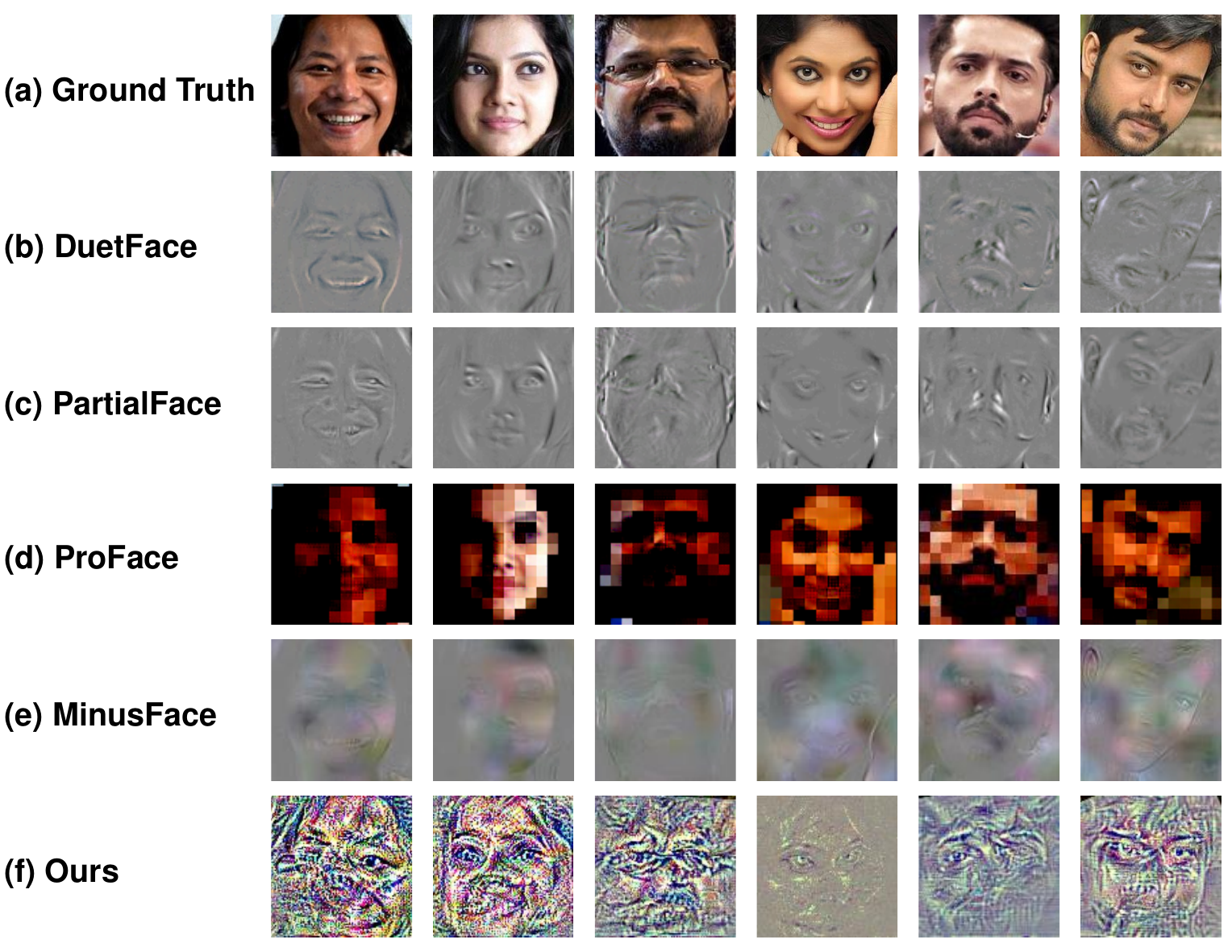} 
    \caption{Additional sample images for the protective $X'$.
}
    \label{fig:anonymization}
\end{figure}

\begin{figure}[htbp] 
    \centering
    \includegraphics[width=1.0\linewidth]{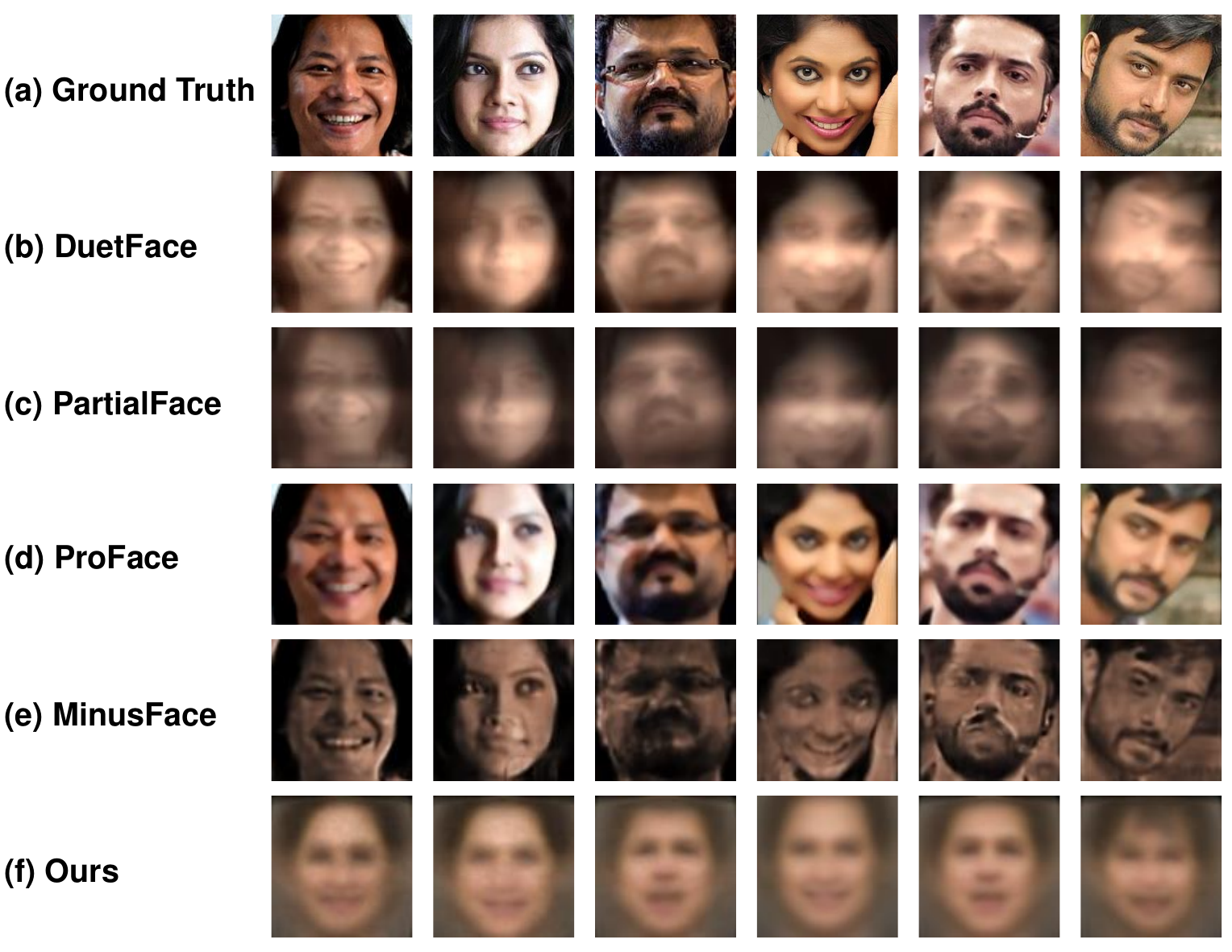} 
    \caption{Additional sample images for the recovery from $X'$.
}
    \label{fig:recovery}
\end{figure}

To demonstrate the generality of our method, we further provide additional examples showcasing the protective representation $X'$ and its recovery, comparing our approach against state-of-the-art methods. Specifically, Fig.~\ref{fig:anonymization} illustrates $X'$ , while Fig.~\ref{fig:recovery} presents the recovery results.

{
    \small
    \bibliographystyle{ieeenat_fullname}
    \bibliography{main}
}
